# Effective faking of verbal deception detection with target-aligned adversarial attacks

Bennett Kleinberg[1,2,*], Riccardo Loconte[1,3], Bruno Verschuere[4]

[1] Tilburg University, Department of Methodology and Statistics, The Netherlands
[2] University College London, Department of Security and Crime Science, UK
[3] IMT School of Advanced Studies, Lucca, Italy
[4] University of Amsterdam, Department of Psychology, The Netherlands

**Abstract**

Background: Deception detection through analysing language is a promising avenue using both human judgments and automated machine learning judgments. For both forms of credibility assessment, automated adversarial attacks that rewrite deceptive statements to appear truthful pose a serious threat.

Methods: We used a dataset of 243 truthful and 262 fabricated autobiographical stories in a deception detection task for humans and machine learning models. A large language model was tasked to rewrite deceptive statements so that they appear truthful. In Study 1, humans who made a deception judgment or used the detailedness heuristic and two machine learning models (a fine-tuned language model and a simple n-gram model) judged original or adversarial modifications of deceptive statements. In Study 2, we manipulated the target alignment of the modifications, i.e. tailoring the attack to whether the statements would be assessed by humans or computer models.

Results: When adversarial modifications were aligned with their target, human ($d$=-0.07 and $d$=-0.04) and machine judgments (51% accuracy) dropped to the chance level. When the attack was not aligned with the target, both human heuristics judgments ($d$=0.30 and $d$=0.36) and machine learning predictions (63-78%) were significantly better than chance.

Conclusions: Easily accessible language models can effectively help anyone fake deception detection efforts both by humans and machine learning models. Robustness against adversarial modifications for humans and machines depends on that target alignment. We close with suggestions on advancing deception research with adversarial attack designs and techniques.

[*] Corresponding author: Bennett Kleinberg, bennett.kleinberg@tilburguniversity.edu

**INTRODUCTION**
Detecting deception matters in daily life and in the courtroom. However, deception detection is also very hard and has explored a range of instruments, methods and approaches (Docan-Morgan, 2019; Granhag et al., 2015). The most often applied means of deception detection is through analysing language, commonly performed by trained humans and practiced by forensic psychologists, police, and business in several countries worldwide. In recent years, several computer-automated approaches have been developed, and AI tools for deception detection are increasingly marketed (e.g., liarliar.ai). But an emerging threat to the validity of deception detection approaches is largely glanced over: can artificial intelligence fool these deception detection techniques? In this paper, we introduce and empirically examine automated adversarial text modifications as threat for verbal deception detection.

**Verbal deception detection**
The central difference between the verbal approach to deception detection and other approaches (e.g., polygraphy, voice stress analysis, behaviour analysis) is that it relies on the content of statements that are either truthful or deceptive (Vrij, 2019; Vrij et al., 2022). Decisions about veracity are based on the information that a participant or suspect provides. On the one hand researchers develop techniques to maximize the information value of the statements (Vrij & Granhag, 2012). On the other hand, there is extensive research trying to improve how the credibility assessment is made; which cues to extract, how to extract them, and how to combine them (Levine, 2014; Verschuere et al., 2023). Investigative interviewing techniques which encourage interviewees to provide a rich statement aim to elicit more diagnostic information than interviews without specific techniques (Mac Giolla & Luke, 2021). Research on the detectability of deception contained in a verbal statement (e.g., transcripts or typed narratives) has consistently shown that humans perform at the chance level when tasked to directly make a deception judgment (e.g., asking "How deceptive is this statement?"; for two meta-analyses, see Hartwig & Bond, 2011, 2014). For a long time, therefore, this task was outsourced to experts trained in extracting cues to deception (Nahari et al., 2019). Recently, however, the notion of poor human deception detection ability by lay people has been challenged with the introduction of a heuristics-based approach (Verschuere et al., 2023): when attention was directed to focusing on a single yet empirically supported cue – the detailedness of a statement (DePaulo et al., 2003; Nahari & Pazuelo, 2015; Taylor et al., 2017; Verschuere et al., 2021; Warmelink et al., 2013) – lay people's ability to detect deception improved (65-70%) compared to a standard deception judgement (50-52%). Importantly, this "use the best, ignore the rest" heuristic asks human judges to assess statements only on that detailedness cue rather than eliciting a deception judgment directly. The rationale of that single-cue approach is to focus humans' attention on what is known to work – detailedness - and then use that judgment to infer deception. In doing so, the heuristics judgment solves the problem of direct deception judgments where humans often rely on the wrong cues (Bogaard et al., 2016; Hartwig & Bond, 2011) or struggle to integrate multiple cues (Hartwig & Bond, 2014). The paper introducing the heuristic (Verschuere et al., 2023) has shown that the approach is robust even when participants knew that the statements they judged can be truthful or deceptive.
 While that promising avenue of research could help improve human deception detection, another research area relies on computational methods for verbal deception detection to increase the scale, reliability and potential performance of detecting deception.

**Computer-automated verbal deception detection**

Similar to human verbal deception detection, the data used to arrive at a computational judgment are verbal statements. The key difference is how the automated approaches make a credibility assessment (Fitzpatrick et al., 2015). Rather than using human judgment, the computational approach relies on machine learning and natural language processing (NLP). First, textual data are quantified (e.g., word frequencies, named entities, psycholinguistic variables, embedding representations) with NLP methods to arrive at a numerical representation of the data that can be used for further statistical analysis. The second step then involves using that numerical representation as input to train a classification algorithm that learns from labelled data (e.g., a logistic regression, support vector machine, neural networks). In the third step, that trained algorithm is assessed on a subset of the data that was not present during the training phase to obtain performance metrics.

Automated verbal deception detection has been used in various contexts (e.g., hotel reviews: Ott et al., 2011, 2013; trial transcripts: Fornaciari & Poesio, 2013; Pérez-Rosas et al., 2015; deceptive intentions: Kleinberg et al., 2018) and typically performs well above the chance level (Constâncio et al., 2023). Advancements in machine learning have also found their way into automated deception research albeit with some lag. For example, large language models such as ChatGPT that were trained on massive corpora of text data and arrived via strictly bottom-up learning at a model of human language, could be promising. These language models have been shown to outperform other computational approaches on the majority of benchmark tasks (e.g., text summarisation, sentiment analysis, and question-answering; Chang et al., 2024). Recent work has used a language model that was refined for deception detection and showed that it outperformed a simpler word frequency-based model – albeit with improvements that were context-dependent and ranged between 5.18 and 23.04 points (Loconte et al., 2023). Others report similar promise but suggest a pronounced truth bias in language models' veracity judgments (Markowitz & Hancock, 2024).

While there is promise in machine learning and NLP methods for deception detection, these techniques also enable a novel and impactful threat to automated and human deception: adversarial attacks.

**Adversarial attacks**

Adversarial machine learning pertains to a subfield of computer science that assesses the robustness of classification models (e.g., image and text classification). An original classification model becomes the target of an attack when an adversarial attack model seeks to modify input in such a way that the target model misclassifies the data. Importantly, the input data modification occurs in such a way that the changes are imperceptible to humans. The classic design of adversarial learning is image recognition. For example, minute changes to individual pixel values can trick the system into misclassifying dog images as cats (Elsayed et al., 2018). Recent work has brought the problem to NLP research (Bartolo et al., 2020; Morris, Lifland, Yoo, et al., 2020).

For example, one study (Mozes, Stenetorp, et al., 2021) used a fine-tuned language model (RoBERTa) that classified positive and negative movie reviews with an accuracy of 94.9%. The adversarial attack first queried the model to understand how it weighed input information to make a positive versus negative movie review classification, then bespoke modifications of movie reviews were created by substituting words with low-frequency synonyms (e.g., "wonderful" became "tremendous"). By modifying the reviews, these adversarial attacks were successful in flipping the target model's prediction and more than halved the classification accuracy (from 94.9% to 40.8%). Adversarial attacks are a central method to assess the robustness of machine learning models but have not yet reached deception research.

For deception detection, adversarial modifications pose a threat when deceptive text data are purposefully modified to appear - to a human or a classifier – as truthful (Fig. 1). Devising adversarial attacks has until recently necessitated advanced machine learning. But the rapid availability of large language models (e.g., ChatGPT, Llama) has lowered the barrier for devising such attacks and would, in principle, enable everyone to instruct a model to modify statements that mislead another human or a deception classifier. Our aim is to test how vulnerable humans and machine learning classifiers are to adversarial modifications created with easy-to-access language models.

*Figure 1*. Schematic overview of an adversarial attack on a deception classifier

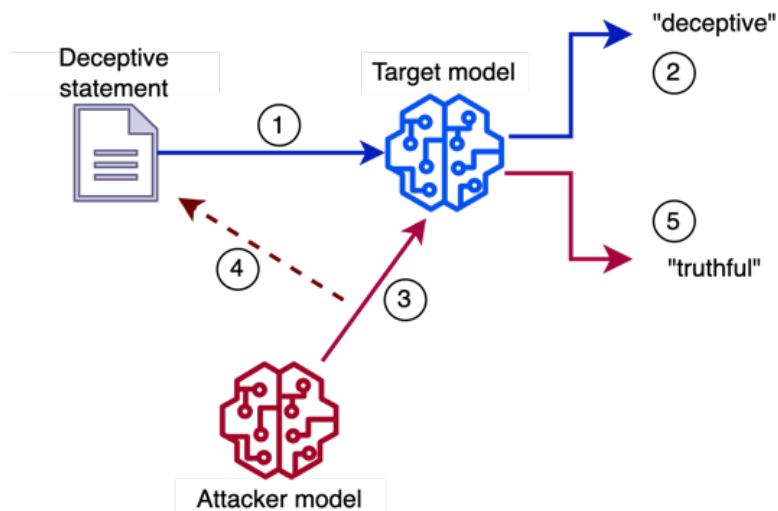

*Note*. In a normal, non-attacked flow, 1) a deceptive statement is passed to a trained model (the target model) and 2) correctly classified as "deceptive". During an adversarial attack, 3) an attacker model exploits the statistical decision-making of the target model and 4) carefully rewrites (e.g., by changing words) the deceptive statement. That perturbation is crafted in such a way that 5) the target model misclassifies the deceptive statement as truthful.

**Aims of this paper**

We test how robust both human judgments and machine learning classifiers are against adversarial modifications for deception detection. We compare deception detection performance on original statements with the performance achieved on adversarial modifications of deceptive statements. These modifications are crafted by a state-of-the-art large language model. In Study 1, we test two modification attack variations. The attack was developed with or without guidance on how to make a statement appear more truthful. We further examine the efficiency of the attack on two human and two automated deception detection approaches. For the human judgments, we compared a direct deception judgment as a control condition with the novel use-the-best heuristic decision-making strategy (Lob et al., 2024; Verschuere et al., 2023). We used machine learning classifiers of two levels of complexity: a fine-tuned language model and a simple word frequency-based model. In Study 2, we expand the adversarial design by devising more targeted automated rewriting modifications (i.e., specifically targeted at a model from study 1 versus targeted at humans) and evaluate them on both humans and a machine learning model.

## STUDY 1

**Method**

Data availability statement

All data, analysis code and LLM-generation code to reproduce the findings of this paper are available at https://osf.io/7qz94/?view_only=73ef083d0a8b44cb9333ab17401d062d.

Truthful and deceptive statements

We used the Hippocorpus dataset (Sap et al., 2020), which contains stories written by a total of 5,047 participants about remembered and imagined events after removal of missing values. Participants were first asked to recall a salient event that they had experienced in the past six months and write a story as well as a 2-3 sentence summary about that event. The summaries were then provided to a new sample of participants who had to imagine that event and write a fabricated story about it.

We evaluated the effect of adversarial modifications on both humans and a deception classifier. That classifier used a train/test split, so to avoid testing in already seen data, we used the test dataset from the classifier (i.e., the subsample of stories not used in the training phase) as the relevant dataset for the textual modifications. That test dataset consisted of 505 stories (262 deceptive, 243 truthful). The training set consisted of the remaining 4542 statements.

Adversarial modifications

The adversarial versions were obtained by prompting a large language model (here: GPT-4-turbo) to rewrite the deceptive statements so that they appear truthful to humans (Table 1). We used two variations of that procedure: in the unguided modification condition, the model was instructed to "*[r]ewrite the following deceptive statement so that it may appear truthful to humans. Keep almost the same length as the original statement*".

In the guided modification condition, we explained that liars try to mislead others by providing details that cannot be easily checked (Hartwig et al., 2007; Nahari et al., 2014a) and directed the model to use unverifiable details with the following instruction: "*We know from research that liars prefer to avoid providing details that can be verified whereas truth-tellers prefer to provide details that can be verified. Verifiable details are (i) activities carried out with identifiable or named persons who the interviewer can consult, (ii) activities that have been witnessed by identifiable or named persons who the interviewer can consult, (iii) activities that the interviewee believes may have been captured on CCTV, and (iv) activities that may have been recorded and documented, such as using debit cards, mobile phones, or computers. Rewrite the following deceptive text by adding UNVERIFIABLE DETAILS so that it may appear truthful to humans. Pay attention to add only* *unverifiable details. Do not add any verifiable detail. Keep almost the same length as the original statement*".

The prompts were sent to the large language model, eliciting completions by the model (here: modifications of an original statement). For the interaction with the LLM via a programmatic API interface, we used the *rgpt3* R package (Kleinberg, 2024). Using the API allows control over more detailed parameters of the requests than is possible in a graphical user interface, such as the widely used ChatGPT interface. One important parameter pertains to the sampling temperature (i.e., the randomness involved in selecting the next tokens in the completion). Since there is no clear evidence yet as to what an adequate sampling temperature value is, we randomly sampled the temperature parameter from a range of 0.01 to 1.00 (see Peereboom et al., 2025). The maximum length of the LLM's response was set to the length of the original statement plus 20 tokens.

Human deception judgment

We recruited participants via the online participant pool Prolific and asked them to assess ten randomly selected statements each. We aimed to have each statement assessed by at least three participants and used the mean of the judgments per statement for further analysis. Upon providing informed consent, participants were provided instructions about their task, which created two judgment conditions.

In the heuristic judgment condition, participants' focus was directed solely on judging the detailedness of the statements using a definition of detailedness as "the degree to which the message includes details such as descriptions of people, places, actions, objects, events and the timing of events; the degree to which the message seemed complete, concrete, striking or rich in details" (Verschuere et al., 2023). In the control condition, participants were asked to make a direct deception judgment assessing each statement on a scale from "0=completely deceptive" to "10=completely truthful".

Within each judgment condition, participants were further allocated to one of the three modification conditions. They either saw the original statements from the Hippocorpus test set, the unguided adversarial modifications or the guided adversarial modifications. In each modification condition, the truthful statements were not modified but were judged anew so that participants were exposed to the same distribution of truthful and deceptive statements in all conditions. This design resulted in ten statements judged by each participant in their respective condition (i.e., a combination of heuristic vs. direct deception judgment and original vs. unguided vs. guided modification). Within that condition combination, the ten statements were randomly sampled regarding statement veracity.

All participants made their judgment on an 11-point scale from 0=not detailed at all [completely deceptive], 5=moderate/neutral, to 10=absolutely detailed [completely truthful]. Participants were aware that the study was about detecting truthful and deceptive stories about recollected or fabricated events from another dataset. After completing the task, participants were debriefed and informed about the adversarial modification that was included in some statements.

Training machine learning models

Two machine-learning approaches were used. First, we retrained a model from previous research (Loconte et al., 2023). Specifically, we fine-tuned a FLAN-T5 base model[1] on the training set for a binary classification task. A 10-fold cross-validation was employed to assess the model's performance. After cross-validation, the model reached an average accuracy of 79.87% (*SD*=2.07), which was sufficient to then retrain it once on the whole training set to obtain a single, final model for use on the test set. Model training was done on Google Colaboratory Pro+ using an NVIDIA A100 Tensor Core GPU.

Second, to include a simpler model, we retrieved a representation of *n*-grams (unigrams, bigrams and trigrams, representing sequences of *n* consecutive words) after removing stop words and punctuation and applying stemming to the words. We further only included words that occur in at least 1% of the statements as a means to reduce the risk of overfitting. From the retained 1621 n-grams as predictors, we removed those with a near-zero variance in the training phase, resulting in a final set of 321 predictors that were used for training the model. The classification algorithm was a support vector machine model trained on 80% of the data with 5-fold cross-validation in the training set. The model was evaluated on the remaining unseen 20% of the data. Since such an n-gram model does not consider the order in which words occur, this is also known as a bag-of-words model. For the model training, we used the *caret* R package (Kuhn, 2008), and natural language processing was conducted using the *quanteda* R package (Benoit et al., 2018).

Table 1. *Verbatim examples (abbreviated) showing two statements in each of the modification conditions.*

| Original | Unguided modification | Guided modification |
|---|---|---|
| About three weeks ago my best friend gets kicked out of her house along with her child and she came to me for help. I gladly accepted her and her child to stay at my home rent free until she is ready to go on. My bestfriend and her husband had a huge falling out. I felt extremely guilty. I allowed her to stay at my house for three weeks now. All she has done for me is take and take from me without ever giving. [...] | About three weeks ago, my best friend was forced to leave her home with her child, and she turned to me for assistance. I welcomed them into my home without expecting rent, hoping to provide support until she could regain her footing. Unfortunately, there have been challenges. Despite her difficult situation, the daily living habits and lack of contribution to household chores have created tension. [...] | About three weeks ago, during a stormy night, my best friend was suddenly kicked out of her house along with her child, and she came to me for help. I gladly accepted her and her child to stay at my home rent-free until she is ready to move on. My best friend and her husband had a huge, emotional falling out that left her devastated. I felt extremely guilty seeing her in that state. I allowed her to stay at my house for three weeks now. During this time, all she has done is take and take from me without ever giving back, but I figured she was deeply hurt and not in a good place emotionally, so I allowed it. [...] |
| One month ago i went to Italy. Me and me wife rented a log cabin. When we got to the cabin we looked at brochures of things to do in the area. at first we went site seeing. after that we saw Dolomite mountain. My wife said we should | Last month, my wife and I traveled to Italy where we rented a charming log cabin. Upon arrival, we explored various local brochures to plan our activities. Initially, we enjoyed some sightseeing around the area. During our explorations, we were | One month ago, I traveled to Italy with my wife. We rented a quaint log cabin nestled in a secluded area, surrounded by lush forests and a serene lake, which shimmered under the moonlight. Upon our arrival, we spent the evening sifting |

---

[1] https://huggingface.co/docs/transformers/model_doc/flan-t5; The parameter configuration for training included a learning rate of 5e-5, a weight decay coefficient of 0.01 and three epochs.

| | | |
|---|---|---|
| go hiking on that mountain range. at first i was skeptical. After more research i decided it would be fun for us to do. We gathered some supplies. Then we started at the bottom of the range. [...] | captivated by the majestic Dolomite Mountains. My wife suggested we hike the mountain range, and though initially hesitant, further research convinced me it would be an enjoyable challenge. We prepared by gathering necessary supplies and commenced our hike at the base of the mountains. [...] | through a collection of old, faded brochures we found in a drawer, which suggested various local sights and activities. Initially, we explored the nearby quaint villages, immersing ourselves in the charming local culture. Later, we caught a glimpse of the majestic Dolomite mountains, which looked even more stunning from a distance. My wife, inspired by the view, suggested we go hiking across those mountains. Initially hesitant, I warmed up to the idea after imagining the breathtaking views we would enjoy from the peaks. We prepared by packing some essentials—snacks we brought from home and warm clothing. [...] |

Analysis plan

We examine the robustness of two deception detection modalities against adversarial rewriting. First, we test how human judges who either use a standard deception judgment or the decision heuristic differentiate between truthful and deceptive statements that were either presented in the original form or rewritten by a language model. Second, we trained deception classifiers on the training set of the data – a simple bag-of-words model and a fine-tuned language model - and assessed their performance on original or rewritten truthful and deceptive statements. For both modalities, a drop in deception detection performance on the rewritten statements would suggest that adversarial modifications were successfully misleading. The main design of the analysis is 2 (Veracity: truthful vs deceptive) by 3 (Modification: original vs unguided vs guided). For human judgments, the manipulation also included the factor Judgment (deception vs heuristic). Unless reported otherwise, we use a significance threshold of *p*<.01 as more conservative criterion than the traditional *p*<.05 to safeguard better against false positive findings (Lakens et al., 2018).

**Results**

Corpus descriptives

The original dataset contained 505 statements (262 deceptive, 243 truthful). For each statement, we obtained six aggregate judgments resulting from the three variations (two adversarial modifications plus the original) and the two human judgment conditions (detailedness vs veracity judgments). Only statements for which we had at least two independent human judgments were included (number of judges per statement: *M*=3.09, *SD*=0.59), resulting in a final dataset of 2998 judged statements. The average word count was 278.02 (*SD*=96.97). There were significant differences in the length of the statements both in original form and in exacerbated form in the adversarial modifications (Table 2).

Table 2. *Average word count per statement veracity and modification.*

|  | Truthful | Deceptive | Cohen's d |
|---|---|---|---|
| Original | 310.66 (98.16) | 274.91 (102.28) | 0.37 [0.14; 0.60] |
| Unguided | 311.36 (98.07) | 206.19 (54.74) | 1.34 [1.08; 1.59] |
| Guided | 311.40 (98.04) | 263.37 (76.20) | 0.56 [0.33; 0.80] |

Note. The subtle differences in the means of the truthful statements result from the decision to only include statements with at least two independent judgments. Note that the truthful statements were identical in each comparison but were judged three times (once in each of the respective deceptive modifications).

Participants in the human judgment task

In total, we recruited *n*=981 participants (54.84% male, 40.67% female, 4.49% data not provided). The average age of the participants was 34.26 years (*SD*=12.80). Each participant was paid GBP 1.50, and the median completion time for judging ten statements was 11 minutes (see Appendix 1 for inclusion criteria per study).

Human judgments

The 2 (Statement Veracity: truthful vs deceptive) by 3 (Statement Modification: unguided modification vs guided modification vs not modified) by 2 (Human Judgment: heuristic vs control) ANOVA revealed a significant main effect of Judgment, *F*(1, 2986)=46.00, *p*<.001, *eta-sq*=0.02, that subsumed under a significant three-way interaction, *F*(2, 2986)=10.99, *p*<.001, *eta-sq*=0.01 (see Table 3). We unpack that effect by Modification and conduct follow-up 2 (Veracity) by 2 (Judgment) ANOVAs for each modification type (Table 3).

For the original texts, a significant Veracity by Judgment interaction, *F*(1, 991)=8.02, *p*=.005, *eta-sq*=0.01 indicated that when using the heuristic, human truthfulness judgments were significantly higher for truthful than deceptive statements, *d*=0.30 [0.07; 0.54] and resulted in a diagnostic value better than chance. Without the heuristic, humans performed at the chance level, *d*=-0.07 [-0.30; 0.16].

For statements that were modified without guidance (unguided modification), we also found evidence for the Veracity by Judgment interaction, *F*(1, 1001)=13.45, *p*<.001, *eta-sq*=0.01. Follow-up tests showed that there was no significant difference between truthful and deceptive statements in the heuristic judgment, *d*=0.21 [-0.02; 0.44]. In contrast, in the control judgment, there was a difference in the opposite direction: deceptive statements were judged as more truthful than truthful statements, *d*=-0.25 [-0.48; -0.02]. The unguided modification rendered the heuristic ineffective and caused confusion in judgments in the control so that deceptive statements were perceived as more truthful than truthful ones.

For the guided modification, there was no significant interaction effect, *F*(1, 994)=6.13, *p*=.013, *eta-sq*=0.01. Consequently, human deception detection ability was at the chance level in both judgment conditions.

Table 3. *Human judgments (M, SD) by statement veracity, judgment type and modification.*

|           |          | Truthful    | Deceptive   | AUC                 | Cohen's d              |
|-----------|----------|-------------|-------------|---------------------|------------------------|
| Control   | Original | 6.00 (1.79) | 6.13 (1.80) | 0.51 [0.45; 0.58]   | -0.07 [-0.30; 0.16]    |
|           | Unguided | 5.88 (1.88) | 6.34 (1.79) | 0.43 [0.37; 0.50]   | -0.25 [-0.48; -0.02]*  |
|           | Guided   | 6.38 (1.73) | 6.50 (1.47) | 0.56 [0.49; 0.63]   | -0.22 [-0.45; -0.01]   |
| Heuristic | Original | 6.73 (1.64) | 6.23 (1.61) | 0.58 [0.51; 0.65]*  | 0.30 [0.07; 0.54]*     |
|           | Unguided | 6.86 (1.88) | 6.55 (1.45) | 0.56 [0.50; 0.63]   | 0.21 [-0.02; 0.44]     |
|           | Guided   | 6.37 (1.55) | 5.99 (1.76) | 0.47 [0.40; 0.54]   | 0.09 [-0.14; 0.32]     |

Note. All confidence intervals in square brackets are 99% confidence intervals. The Cohen's d value is the effect size of the Veracity-by-Judgment interaction. *=p<.01

Machine learning deception classification

Both deception classifiers - the fine-tune language model and the bag-of-words model - performed significantly better than the chance level (Table 4). While the bag-of-words model showed accuracies that were 7-14 percentage points lower (accuracies: 0.64-0.65) than those of the fine-tuned language model (accuracies: 0.71-0.79), the variability across statement modification conditions was lower than for the language model. In particular, the unguided modification resulted in a drop of seven percentage points for the language model (0.78 to 0.71) but did not affect the bag-of-words model (0.64 remained 0.64). Differences between the two models can largely be attributed to the bag-of-words model's poor recall values (i.e., detection rate) for truthful statements of 0.46.

Table 4. *Classification performance on the original and modified statements for both machine learning models.*

|                    |          | Accuracy | AUC               | Precision (truthful) | Recall (truthful) | Precision (deceptive) | Recall (deceptive) |
|--------------------|----------|----------|-------------------|----------------------|-------------------|-----------------------|--------------------|
| Bag-of-words model | Original | 0.64     | 0.67 [0.61; 0.74] | 0.68                 | 0.46              | 0.62                  | 0.80               |
|                    | Unguided | 0.64     | 0.68 [0.62; 0.74] | 0.70                 | 0.46              | 0.62                  | 0.81               |
|                    | Guided   | 0.65     | 0.69 [0.63; 0.75] | 0.71                 | 0.46              | 0.63                  | 0.82               |
| Fine-tuned LM      | Original | 0.78     | 0.79 [0.73; 0.84] | 0.78                 | 0.76              | 0.78                  | 0.80               |
|                    | Unguided | 0.71     | 0.75 [0.69; 0.80] | 0.67                 | 0.77              | 0.75                  | 0.65               |
|                    | Guided   | 0.77     | 0.85 [0.80; 0.89] | 0.75                 | 0.77              | 0.78                  | 0.77               |

*Note.* The models were not optimised for recall (i.e., how many statements of a class are detected) or precision (i.e., how accurate the predictions for a class are) of a particular class but for overall AUC.

Discussion Study 1

Our findings suggest that humans used the use-the-best heuristic effectively to improve their deception detection ability – albeit to a smaller effect size than originally reported. However, when statements were modified, the heuristic became ineffective, bringing judgments back to the chance level. In contrast, when statements are judged by machine learning classifiers, the drop in deception detection performance as a result of adversarial modifications was less substantial and remained above the chance level. A simple word frequency-based model performed 14 percentage points below the larger, fine-tuned language model but was more robust against adversarial modifications.

The modifications in Study 1 targeted a presumed mechanism in human deception (i.e., providing unverifiable details to appear credible) and may thus have put humans at an unfair disadvantage. To test whether the target of the modifications affects the judgment robustness, we manipulated the target in Study 2. Specifically, we devised modifications targeted either at humans or at the bag-of-words model from Study 1. If target alignment played a role, we would expect modifications to be more effective when modification target and actual judgment modality are the same (human target and human judgment; model target and model judgment) than if they are divergent.

**STUDY 2**

**Method**

Targeted adversarial modifications

The key difference to Study 1 was how we devised the adversarial modifications. We used the same procedure with the key differences that the attack was now tailored to either humans or the bag-of-words model from Study 1. The bag-of-words model was chosen because, in contrast to the fine-tuned LM, we can trace which features it used and provide - at least superficially - some explainability for the modification attack. Also, Study1 shows that this bag-of-words model was most resistant to adversarial attacks as neither the guided not he unguided attacked lowered its accuracy. Two other changes were made to the design: i) since running Study 1, a more capable LLM has been released (GPT-4o) which we used to reflect the upper boundary of what we can expect an LLM to do under our research design; ii) we chose a recommended default sampling temperature for model of 0.7; and iii) we changed the modification instructions to specifically target humans or the bag-of-words model from Study 1.

For Study 2, the original statements were rewritten again, this time tailored either to fool humans or to fool the classifier that proved to be most robust in Study 1. For the human-targeted instructions, the prompt to the LLM read: *"Below is a deceptive statement that was written by a human. Your task is to rewrite that statement in such a way that it appears more truthful to a human. The humans who read and assess the statement will be focusing on the detailedness and use detailedness as an indicator of truthfulness. For them, more details suggest a higher probability of truthfulness. Your task is to rewrite the statement so that it appears more truthful to humans. Finish your statement with a complete sentence and adhere to a similar length as the original statement when you rewrite it."*

The model-targeted instructions, in turn, included details (highlighted below in bold) about the kind of model, the model's confidence in the current statement being truthful (based on the predictions on the unmodified statements from Study 1), and the top ten features used by the model. The prompt read: *"[…] Your task is to rewrite that statement in such a way that it appears more truthful to **a machine learning text classifier**. That classifier is a **bag-of-words model** with the following properties: The **most important ngrams (in stemmed format) for the model to make a decision were: ago, year, event, recent, month_ago, memor, day, also, week_ago, last**. For the statement below, the model predicted that the probability of this statement being truthful is [XY]%. Your task is to increase that probability by modifying the statement. […]"*. Each deceptive statement was modified in both target versions. These were then joined with the unmodified truthful statements for the human and machine learning judgments.

Human judgments

The procedure followed the one from Study 1. This time, all participants used the detailedness heuristic, dropping direct veracity judgments where humans performed at chance level even on the unaltered statements. All participants judged 10 statements. All modified statements were judged in two separate tasks, so that for each modification target, truthful and deceptive statements were presented in a random selection. The original, unmodified deceptive statements were not judged.

Machine learning models
We used the same bag-of-words model as in Study 1 (i.e., there was no additional training of the model with new data).

Analysis plan
For human judgments, we use a 2 (Statement Veracity: truthful vs deceptive) by 2 (Statement Modification: human-target vs model-targeted) ANOVA on the average human judgment per statement. Machine learning performance will be compared by accuracy and AUCs.

**Results**
Participants in the human judgment task
The human judgments were crowdsourced from *n*=302 participants with an average age of 34.68 years (*SD*=11.97), of whom 52.31% were male, 46.36% female and 1.33% without gender data. Each participant was paid GBP 1.50 for a median task completion time of 11 minutes.

Corpus descriptives
For statements modified for humans as targets, the truthful ones (*M*=359.82, *SD*=194.07) were significantly longer than deceptive ones (*M*=241.84, *SD*=64.66), *d*=0.83 [0.59; 1.03]. This effect was somewhat more pronounced when the target was the machine learning model (truthful: *M*=358.09, *SD*=192.41; deceptive: *M*=214.94, *SD*=92.97), *d*=0.96 [0.72; 1.21]. As in Study 1, these length differences are larger than in the unmodified statements. The human-targeted deceptive modifications were also longer than the model-targeted ones, *d*=0.34 [0.11; 0.56]. Each statement was rated, on average, 3.04 times (SD=0.52).

Human judgments
The 2 (Veracity) by 2 (Modification target) ANOVA indicated a significant main effect of Veracity, $F(1, 984)=6.77$, *p*=.009, *eta-sq*=0.01. The main effect of modification was not significant, $F(1, 984)=1.38$, *p*=.240, *eta-sq*=0.01. The significant veracity by modification interaction, $F(1, 984)=9.97$, *p*=.002, *eta-sq*=0.01, indicated that when the modification was targeted at humans, the deception ability of humans was at the chance level, *d*=-0.04 [-0.20; 0.27]. In contrast, when the modification targeted the machine learning model, humans' detection ability was significantly above the chance level, *d*=0.36 [0.13; 0.60] (Table 5). These findings suggest that targeted modifications at humans are indeed highly effective. The findings also suggest that when the attack is not targeting a presumed human decision-making mechanism, the effectiveness of the adversarial modification vanishes and the heuristic results in above chance level human judgments.

Table 5. *Human judgments (M, SD) by statement veracity and modification target*

|  | Truthful | Deceptive | AUC | Cohen's d |
|---|---|---|---|---|
| Target human | 6.87 (1.51) | 6.93 (1.43) | 0.50 [0.43; 0.56] | -0.04 [-0.20; 0.27] |
| Target machine learning model | 7.07 (1.53) | 6.53 (1.44) | 0.61 [0.54; 0.68] | 0.36 [0.13; 0.60]* |

Note. All confidence intervals in square brackets are 99% confidence intervals. The Cohen's d value is the effect size of the Veracity-by-Judgment interaction. *=p<.01

Machine learning deception classification

The classification results (Table 6) echo the findings from the human judgments. When target and judgment modality were aligned, the modification was effective, rendering the classification at the chance level (accuracy=0.51). When there was divergence in the target and modality, there was no such effect, and the performance remained above the chance level (accuracy=0.63) essentially replicating the performance on the unmodified statements from Study 1 (accuracy: 0.64).

Table 6. *Classification performance of the bag-of-words model from Study 1 by modification target*

|  | Accuracy | AUC | Precision (truthful) | Recall (truthful) | Precision (deceptive) | Recall (deceptive) |
|---|---|---|---|---|---|---|
| Target human | 0.63 | 0.65 [0.59; 0.71] | 0.64 | 0.47 | 0.62 | 0.77 |
| Target model | 0.51 | 0.48 [0.41; 0.55] | 0.49 | 0.46 | 0.53 | 0.55 |

Semantic similarity constraints

In adversarial machine learning on textual data, a commonly used constrained is that the adversarial modification must preserve the meaning of the original statement. This is typically measured computationally as the cosine similarity between embeddings representations of the original and modified text (Morris, Lifland, Lanchantin, et al., 2020). Embeddings capture semantic information in textual data (Mikolov et al., 2013; Pennington et al., 2014). The cosine similarity between two embeddings representations ranges from 0.00 (no similarity) to +1.00 (identical). If the similarity is too low, the modifications can be deemed invalid, so it merits attention to test whether this criterion is met and how it affects the modification effectiveness.

For the modifications in Study 2, for the human-targeted modifications, 97.69% had a similarity with the original more than 0.80 and 65.77% a similarity of 0.90 or higher. These values were higher for the model-targeted modification: 100% and 90.77% had a similarity to the original of 0.80 and 0.90, respectively. These observations manifest in an average similarity for human-targeted modifications (*M*=0.91, *SD*=0.04) that was significantly lower than those for the model-targeted modifications (*M*=0.95, *SD*=0.03), *d*=0.95 [0.72; 1.19]. These findings suggest that the deviations from the original were rather minimal but still somewhat more pronounced when the target were humans, which might be an artefact of the difference in task instructions (i.e., targeting the detailedness for humans versus modifying individual n-grams for the model-targeted attack).

Human evaluation of adversarial modifications

In addition to automated semantic similarity metrics, human evaluations can help assess whether an adversarial modification did not deviate too much from its original statement. The original dataset (Sap et al., 2020) used summaries of truthful events as the basis for deceptive statements, which offered a means for us to assess how adequate a statement was given the provided summary. We recruited a new sample of $n$=244 participants ($M_{age}$=40.67 years, $SD_{age}$=4.16, 52.87% male) via Prolific (remuneration: GBP 1.50; median completion time: 13 minutes) to assess how realistic a statement (original deceptive, human-targeted modification, or model-targeted modification) was given the summary it was based on. Specifically, each participant saw a random selection of ten summary-statement pairs and indicated their agreement on a 5-point scale (1=strongly disagree, 5=strongly agree) to the question "this statement follows realistically from the provided summary" (see Appendix Table 2). Each summary-statement pair was assessed, on average, by 3.15 humans (SD=0.91), and human assessments were averaged per summary-statement pair. If the adversarial modifications of the original deceptive statements had deviated substantially from the original, we would expect lower agreement scores in the human assessment for the two (human-targeted and model-targeted) modification conditions than for the original condition.

The data suggest that there was no difference between the true summary-statement pairs (i.e., the ones that the original deceptive texts were based on, $M$=3.89, $SD$=0.76) or the human-targeted ($M$=3.83, $SD$=0.74) or model-targeted modifications ($M$=3.85, $SD$=0.74), $F(2, 693)$=0.36, $p$=.697, eta-sq=0.00. A Bayesian ANOVA suggested that there was substantial evidence in favour of an intercept-only model (i.e., evidence for no effect), $BF$=43.11. These data indicate that the modifications were as realistic as the original statements when assessed against the source summary, thereby supporting the semantic similarity findings that the modifications were indeed valid adversarial modifications without substantial semantic deviation.

Exploring vocabulary complexity differences between modification targets

To understand better the potential mechanisms through which the modifications achieve their goal, we explored a candidate linguistic variable reported elsewhere when a language model were tasked to mislead humans (Kleinberg et al., 2024). In that study, when instructed to appear human, a language model generated texts with a markedly simpler vocabulary compared to a condition without specific instructions to mislead readers. We measured the vocabulary complexity by comparing the average word frequency ranks obtained from a ranked list of the most common 10,000 words in Google's Trillion Word Corpus (https://github.com/first20hours/google-10000-english). Lower ranks imply a more common and hence less complex vocabulary.

There was a marked effect in vocabulary complexity. The average word frequency rank was lower when the target was a model ($M$=1010.77, $SD$=199.52) than when humans were the target ($M$=1188.52, $SD$=177.53), $d$=0.94 [0.70; 1.18]. That effect suggests that the vocabulary used by the adversarial modification to mislead humans was significantly more complex.

Discussion Study 2

In the second experiment, we assessed whether alignment between target and modality mattered. The findings showed that when these were aligned, the adversarial modification is significantly more effective, both on humans and on a machine learning model. Conversely, when the judgment modality and the target were misaligned, the modifications were ineffective. Under misalignment, humans were able to apply the detailedness heuristic ($d$=0.36) to a similar degree as in the unmodified statements in Study 1 ($d$=0.30). Similarly, the machine learning model's accuracy (0.63) was close to that on the unmodified statements from Study 1 (0.64). Evaluation of the adversarial modifications with automated similarity metrics and human-rated summary-statement matches suggested that the modifications remained sufficiently similar to the original and thus constitute valid adversarial modifications.

## GENERAL DISCUSSION

The availability and capabilities of large language models present not just a promise but also a challenge for verbal deception detection. Language model technology provides easy access to AI tools that may help to deliberately deceive in language. This work provided a first glimpse at this novel challenge known as adversarial attacks and examined the vulnerability of humans and automated classifiers to targeted efforts that invoke misclassifications of deceptive statements as truthful.

Human judgment
When humans judged the unmodified statements, we found evidence for the use-the-best heuristic in both studies. Focusing participants on the detailedness of the statements resulted in significant truth-lie differences in human judgment, while intuitive judgment did not. Compared to other studies on that heuristic, the effect was substantially smaller here ($d$=0.30-0.36, compared to $d$=0.97 in Study 5 of Verschuere et al., 2023). While it is not uncommon to find smaller effects in (partial) replication efforts (Camerer et al., 2018), it merits attention that the heuristic did not work to the same degree on the dataset we used. As reported in large-scale replication efforts (Camerer et al., 2018; Open Science Collaboration, 2015), there are various potential reasons for the often substantially small effect sizes of replication studies. A potential explanation for our finding could be the context of deception. Large effect sizes in favour of the heuristics approach stem from judgments of lies and truths about campus activities that participants actively experienced (Verschuere et al., 2023). In contrast, we used a dataset of truthful versus fabricated autobiographical stories that were – in the truthful condition - recalled from up to six months ago and then fabricated by others based on a summary of that event. This may have made the deceptive autobiographical statements more similar to the truthful statements. Furthermore, the elapsed time between the event and the statement may have diminished the detailedness of the truthful statements in our dataset (Masip et al., 2005; Nahari et al., 2014a). Judging deception in our context would then have been more difficult due to the less pronounced truth-lie differences in detailedness, for example. Another explanation could be the nature of statement elicitation, which included interviewing techniques such as probing for detailedness or rapport building aimed at maximizing lie-truth differences in detailedness (Vrij et al., 2014) in the original heuristics studies.

Importantly, for our key research question, we saw that when adversarial modifications of deceptive statements were presented, the human performance in detecting deception dropped to the chance level in the heuristics condition. A refined design in Study 2 provided more nuance and indicated that this is only the case when the modification directly targets human judgment strategies by increasing the detailedness of a statement. When the adversarial modification was not targeted at humans, the heuristic remained effective. The overall finding that – under the right alignment condition – modifications are effective in moving deception towards truthfulness can be interpreted against the background of True Default Theory (Levine, 2014): the goal of the attacker model (i.e., making deceptive statements appear more truthful) could have been achieved by capitalising on the truth bias. With a human tendency to judge most statements as truthful, the task for the attack model could potentially have been achieved by making modifications to the original text that invoked the truth bias. If this were the case, we would expect in future work that the current attack design of increasing truthfulness is easier to achieve than modifying truthful statements so that they appear more deceptive. There is evidence that the truth bias is even more pronounced in AI models than humans (Markowitz & Hancock, 2024; although in our study, the machine learning model did not exhibit truth bias; but also differed from the LLM classification approach in that paper), which holds specific predictions that could be tested in future work. For both humans and AI models as targets, rewriting truthful statements to appear deceptive would be expected to be more difficult than the opposite, since accusatory judgements are less common and typically avoided (Levine, 2014, 2020b; Von Schenk et al., 2024). TDT further holds that people abandon their truth default state when suspicion is raised by triggers, such as believing that someone may have a reason to lie or displays behavioural cues believed to be associated with lying (Levine, 2014, 2020b). An adversarial attack could exploit these human judgment tendencies, for example, by removing or attenuating these triggers, which would be expected to make deceptive statements be judged by humans as more deceptive, on average. Conversely, subtly inserting such triggers would be expected to lead to truths being judged as lies. What is needed for testing these predictions is future work that examines, in-depth, the strategies used to achieve the adversarial goal and compares human attackers and AI models in their strategies. It might be possible that both employ the same strategies including those expected from deception theory (e.g., adding/removing unverifiable details: Nahari et al., 2014a; adding/removing contradictions: Masip et al., 2018) or that an AI model resorts to entirely novel and unexpected strategies, which could in turn inform deception theory by inferring reliance on specific textual characteristics from the change that their modification has on human judgments (see also "Looking ahead" below).

Machine learning models

When the adversarial modifications were used for automated deception classification with machine learning models, there was clear evidence of target alignment. When the modification did not target a machine learning model (Study 1 and one condition in Study 2), there was no substantial effect. Both classifiers remained well above the chance level (64% and 78%). There was a drop in performance for the language model when modifications were unguided (i.e., not following the inverse verifiability approach logic) from 78% to 71%. It is noteworthy that such a drop was not observed for a simpler n-gram model, which remained consistently between 64-65%. These findings are somewhat surprising, considering previous work that suggests that practically all text classification models are vulnerable when presented with adversarial examples (Alzantot et al., 2018; Bartolo et al., 2020; Morris, Lifland, Lanchantin, et al., 2020; Mozes, Stenetorp, et al., 2021). Moreover, the model that relied solely on the term frequency (n-grams) was less susceptible than the more complex one that uses an entire language model and embedding representations. However, when we devised adversarial modifications specifically targeted at the n-gram model, we observe a similar back-to-chance-level effect as in humans when they were targeted (i.e., an accuracy of 0.51 for aligned modifications and 0.63 for misaligned modifications).

Aligned grey-box adversarial modifications

Our data provide two sides of the same coin. On the one hand, when an adversarial attack (here: textual modifications) is aligned with the target, a simple language model query is sufficient to render human and machine learning judgments ineffective with significant drops of performance. In machine learning terminology, these attacks resemble white-box attacks (Papernot et al., 2016) where details about the target model's workings are available to the attacker. We also provided the attacker (here: a language model tasked with re-writing text to appear more truthful) information about the targets: for humans, we revealed that they pay attention to the detailedness and for the target model, we provided basic model information, class probabilities and the most diagnostic features. In strict white-box attacks, the attacker has access to the gradient and receives constant feedback on the effect of input modifications. In contrast, black-box attacks provide no information about the target. Our approach can thus best be described as a grey-box attack.

On the other hand, as strong as aligned attacks were on humans and a machine learning model, as ineffective were they when they were misaligned with the target. In a deliberately misaligned scenario, none of the modifications had the desired effect, suggesting that, while easily crafted with accessible language models, an attack needs to be aligned to the target model. Without knowledge of the target model, attacks become less effective.

That adversarial attacks utilise word frequencies as a mechanism for altering statements without affecting the meaning is reported elsewhere (Hauser et al., 2023; Mozes, Stenetorp, et al., 2021) and is likely due to the discrete nature of the adversarial modification problem with textual data. When a text should be modified without changing its meaning, substituting words with synonyms that are inevitably often lower in frequency, remains one of few options. Interestingly, we know that when humans are tasked to play the adversary (i.e., humans rewrite text data to mislead a model), they resort to that high-to-low frequency replacement strategy much less often than automated attacks (Mozes et al., 2022) and are markedly more efficient in achieving their goal than automated attacks. Future work could assess how humans attack a deception classifier and human judgments to shed light on i) whether humans are better than an automated attack in misleading others in trusting a statement and ii) how humans think another human and a model rely on textual information for a deception judgment.

Limitations

The focal point of this work is adversarial modifications. We operationalised attacks through an easily available large language model that possesses text manipulation capabilities that far exceed those of other non-generative machine learning models (Chang et al., 2024). The limitation inherent to that approach is that the modifications are much less constrained compared to a model built solely for substituting individual words. Consequently, the validity of modifications is less clear. On the one hand, allowing an attack to paraphrase an input sequence resolves some of the known issues of word-level substitution attacks where grammatical issues often arise due to the difficulty of replacing words with synonyms that also maintain the entire grammatical structure of the statement (e.g., replacing "an error" with "a<u>n</u> mistake", Morris, Lifland, Lanchantin, et al., 2020). At the same time, however, paraphrasing might result in more substantial semantic deviation from the original statement. For the paraphrasing attack reported here with the aim of making deceptive statements appear more truthful, challenge is thus to sense whether the original, deceptive statement was only modified in such a way that the overall meaning of the original was preserved. This is an important property of valid adversarial modification that was assessed in two ways: computational semantic similarity metrics and manual, human assessment. Both approaches suggested that the modified statements were sufficiently similar to the originals. An exciting avenue for future work is to identify the strategies through which a paraphrasing attack seeks to modify a deceptive statement so is perceived as more truthful. In principle, an attack model could achieve this in multiple ways, including applying verbal deception theory (e.g., adding non-verifiable details), but also by making a statement more ambiguous and harder to judge. In both cases, detailed examination of the strategies can inform deception research.

Another angle worth exploring is whether the modifications were good enough. Adversarial machine learning typically involves training an attack model on the weights of the target model (in the case of white box attacks) or relies on significant amounts of trial-and-error with constant feedback from the model, and allows for thousands of iterations until an effective perturbation is found. The approach in our current study, in contrast, relied on a single-shot modification without any training (i.e., a *single-shot grey-box* attack). Future work on adversarial attacks in verbal deception research could compare existing attack frameworks (Morris, Lifland, Yoo, et al., 2020) and test a range of attack types (e.g., strictly word-level substitutions versus character-level attacks).

Further, none of the targets in our study were warned of potential faking attempts through adversarial modifications. Possibly, telling human participants and implementing a filtering mechanism for machine learning models about faking attempts, could attenuate attack effectiveness. Future work could test this experimentally and further our understanding of robustness against targeted modification attacks.

Lastly, the rapid evolution of generative language models implies that the findings presented here are a snapshot of the capabilities of LLMs at the time of data collection. It is possible that future versions of models will be even better at modifying statements towards a specific target. Importantly, models to detect deception automatically might also evolve (e.g., by incorporating knowledge into models), leaving human assessors particularly vulnerable when the sophistication of modification increases. Efforts on human deception detection (Levine, 2014, 2020a; Verschuere et al., 2023) as well as hybrid human-machine deception detection (Kleinberg & Verschuere, 2021; Von Schenk et al., 2024) will thus be particularly pressing avenues in the future.

Looking ahead: Understanding deception with adversarial attacks

Aside from the needed assessment of the robustness of deception detection approaches, the introduction of adversarial attacks into deception research holds exciting potential. Early research on human adversarial attacks (Bartolo et al., 2020; Mozes, Bartolo, et al., 2021; Mozes et al., 2022), showed that humans craft adversarial examples rather differently than automated attacks. Humans are more efficient than the best attack models (e.g., 11 attempts when querying a model interactively versus 140,000 queries for some automated attacks), better at preserving semantics in their attack, and have a more targeted approach to selecting candidate words for replacement. These are in line with expectations that humans can rely on a nuanced language understanding, while automated models need to rely on scale and iterative trial-and-error more often (Mozes, Bartolo, et al., 2021).

What these studies suggest is that putting humans in the shoes of an adversary tasked to mislead a model can results in fundamentally new insights about how humans *think* that a model makes its decision. These insights are derived indirectly from contrasting the strategies employed by automated and human attacks. For deception research, this avenue could break new ground. Not only can we adopt the human adversary design to understand how humans think a deception classifier works, which can inform research on explanability needs (Oswald et al., 2018) and human-AI interaction (Kleinberg & Verschuere, 2021; Von Schenk et al., 2024). More substantively, we can expand this research to a human-on-human attack design: by letting humans attack human judgments (i.e., rewriting text to mislead a human in their credibility assessment), we can infer how an attacking human thinks another human – and oneself by proxy - will arrive at a credibility assessment. Contrasting that with machine-on-human and human-on-machine attacks will enable us to understand different implicit representations held by humans and AI models about deception and could potentially prove a fruitful avenue to inform deception theory.

Implications for legal and criminological psychology

This paper concerned a method from machine learning research applied to automated verbal deception detection. We showed that adversarial machine learning can be employed to make deceptive statements appear more truthful. Although this technique is rather novel in legal psychological research, the consequences are already far-reaching. For example, the assessment of a statement's credibility is crucial in sexual abuse cases (Amado et al., 2016; Griesel et al., 2013), used to decide on insurance claims (Harvey et al., 2017; Leal et al., 2015), and essential in every day police interviewing situations (Granhag, 2024; Vrij et al., 2014). There is already evidence that deceptive individuals can be coached – to an extent – to appear truthful (Vrij et al., 2002). With the advent of advancing artificial intelligence technology, every individual has got access to the tools needed to modify statements with this target in mind. Future research both on automated as well as human approaches to verbal deception addiction should consider the changed backdrop that ease of access to AI technology has created. Several applications of verbal deception detection will face the challenge of automated, adversarial approaches. Moreover, while modifications towards more truthfulness are challenging for deception detection, the opposite – modifying truthful statements to be perceived as deceptive – might be concerning for its impact on trust and social cohesion (Stavrova & Ehlebracht, 2016). Inevitably, this brings with it questions about the ethical implications of research on this matter.

Our work is the first to show how large language models can effectively mislead verbal deception detection with targeted attacks. Others have indicated an extensive truth bias in large language models and the risk that comes with it (Markowitz & Hancock, 2024). Our work suggests that the generative abilities of large language models pose an additional substantial threat to verbal deception detection. We believe the research community needs to be aware of this. Ideally, with awareness of this threat, deception detection efforts improve, be it by building better models including those that detect modifications, devising human deception detection approaches that are robust by design (e.g., similar to how the Verifiability Approach continues to work, even when its rationale is explicitly told to truth-tellers and liars because truth-tellers understand and are willing and able to provide verifiable details, and liars risk being caught when providing details that can be verified; Nahari et al., 2014b; Verschuere et al 2021), or by harnessing the finding that only when an attacker is aware of the target, can an adversarial attack render deception detection approaches ineffective.

There is also promise in adversarial modifications that verbal deception research could exploit. Most importantly, what is known as adversarial training (Bělohlávek et al., 2018), could be part of the solution to the relatively small sample sizes in deception research (Kleinberg et al., 2019) when compared to the scale of machine learning research. In adversarial training for a machine learning classification problem (e.g., deceptive vs truthful statements), an attack model is used to craft adversarial modifications for either class of statements. These modifications then inherently contain slight variations and added complexities of the original statement. Using these as additional examples to train machine on a classifier has been shown to make classifiers more robust (Morris, Lifland, Yoo, et al., 2020; Yoo & Qi, 2021). Concretely, automated verbal deception detection research could build adversarial training approaches to increase dataset sizes by orders of magnitude (e.g., creating 1,000 modifications for each deceptive and truthful statement) and then train models on a notably larger dataset. Such training procedures on larger and more varied datasets might improve automated deception detection. Taken together, adversarial attacks are a problem and opportunity for deception research.

**Conclusions**

Deception detection remains a hard problem but recent advances in human heuristics-based judgment and in automated verbal deception detection are promising. However, when aligned automated adversarial attacks seek to rewrite deceptive statements to appear truthful, human judgments and machine learning classifications dropped to the chance level. Incorporating adversarial attacks in deception research may offer exciting potential to improve deception detection robustness and understand human and machine-held representations of deception.

# Appendix

*Appendix Table 1. Participant inclusion criteria on Prolific per study.*

| Study | Task | Participant inclusion |
|---|---|---|
| Study 1 | Deception judgment (heuristic vs direct deception) | - All countries allowed<br>- First language: English |
| Study 2 | Deception judgment (heuristic-only) | - All countries allowed<br>- First language: English<br>- Not participated in Study 1 |
| Evaluation (Study 2) | Summary-statement rating | - All countries allowed<br>- First language: English<br>- Not participated in Study 1-2 |

*Appendix Table 2. Verbatim examples (unabbreviated) of summary-statement pairs per condition.*

| Summary | Original deceptive statement | Huma-targeted modification | Model-targeted modification |
|---|---|---|---|
| We lost our beloved dog to metastatic liver cancer after several months of trying to do everything we could to make him better. We took him to the vet together and we were with him, holding him, when he breathed his last. We will miss him forever. | 2 months ago my heart was shattered, and I haven't been the same since. That's when we lost Buddy to liver cancer. Looking back he hadn't been the same since the end of last year, but I didn't see it at the time. It started with throwing up more, and losing a bit of weight. I just assumed it was part of him getting older, nothing more. Then one day I noticed he was laying down in an awkward position, and wouldn't sleep. I took him to the vet, and when they called me in to that special room where they give you terrible news, time just stopped. He had advanced liver cancer, and there was not much hope. I fell apart right then and there - Buddy was my entire world. He had seen us through good times and bad, and was our ever constant companion. I called Josh, and told him to come home through the sobs. The vet gave us our options that included an aggressive dialysis schedule, or comfort measures. I couldn't bear the thought of not having Buddy, so we went with the aggressive dialysis schedule. I took him 3 times a week to the animal hospital, and would sit with him for hours | Two months ago, my life changed dramatically when Buddy, our beloved pet, passed away due to liver cancer. Reflecting on the past year, I now realize the signs were there, but at the time, I attributed his increased vomiting and slight weight loss to the natural aging process. One day, I noticed Buddy lying in an unusual position and unable to sleep, prompting a visit to the vet. In the dreaded room where bad news is delivered, I was informed that Buddy had advanced liver cancer with little hope for recovery. I was devastated; Buddy was an integral part of our lives, always by our side through thick and thin. Overcome with grief, I called Josh, urging him to come home immediately. The vet presented us with options: an aggressive dialysis regimen or comfort care. Desperate to keep Buddy with us, we chose dialysis, bringing him to the animal hospital three times a week and staying by his side for hours each visit. Despite our efforts, it became clear over the following months that his condition was worsening. Josh gently persuaded me that it was time | Recent months have been incredibly difficult for me since losing Buddy to liver cancer. Reflecting on it now, I see that he hadn't been himself since the end of last year. Initially, he started vomiting more and losing weight, which I mistakenly attributed to aging. Then, just weeks ago, he began lying in awkward positions and wouldn't sleep, prompting me to take him to the vet. There, in that dreaded room for delivering bad news, I learned he had advanced liver cancer with little hope. My world crumbled at that moment because Buddy was everything to me. He had been our constant companion through both joyful and challenging times. I called Josh, sobbing, asking him to come home. The vet presented us with options: aggressive dialysis or comfort care. Unable to imagine life without Buddy, we chose aggressive dialysis, taking him to the animal hospital thrice a week. Despite holding onto hope for months, it became clear he was still declining. Josh finally convinced me that it was time to let Buddy go. The vet came to our home so Buddy could |
| | before taking him home. I held out hope for a few months, but it became apparent that he was still deteriorating. Josh convinced me that it was time to let Buddy go. The vet came to our house so Buddy would be kept comfortable. We stroked his fur and cried together as he finally was at peace. | to let Buddy go peacefully. The vet came to our home, ensuring Buddy's comfort as we stroked his fur and tearfully said goodbye, finally allowing him to rest in peace. | be comfortable, and we tearfully stroked his fur as he found peace. |
| My mother almost died getting a pacemaker put in her heart. She suffered from a pulmonary edema right before the surgery. She was able to recover in the end. | 2 months ago, one of the worst things I could think of almost happened to me. My mom was going into surgery, she had a weak heart-- Pulmonary Edema. I was extremely concerned, but I had assuaged those fears by believing in the skill of the doctors working with her. I guess that wasn't enough, though. People make mistakes. The surgery went wrong, and my mom was put into a coma in order for doctors to go from extending her lifespan to saving her life. She was in it for a few days, and I was in pieces. My mother was dying, and there was nothing I could do about it. I waited by her side for days. I couldn't eat, I couldn't sleep, I just wanted her to be ok. My dad was worse, though. He couldn't even come in without bawling. He was a strong man, and this was the first time I'd ever seen him just lose it. He was crying so much, I just wanted to hug him and tell him everything would be ok. I was more convincing myself though. Eventually she pulled through, and it was the happiest I ever was. I'm going to make sure she knows how much we love her from now on. | Two months ago, I faced one of the most terrifying experiences of my life when my mom underwent surgery due to her weak heart condition, specifically Pulmonary Edema. I was filled with anxiety but tried to calm myself by trusting in the expertise of her medical team. Unfortunately, even skilled professionals can have setbacks, and the surgery encountered complications. As a result, my mom was placed in a medically induced coma, transitioning the doctors' focus from extending her life to urgently saving it. She remained in the coma for several days, during which I felt utterly helpless and shattered. I stayed by her side, unable to eat or sleep, consumed by the hope that she would recover. My dad, usually a pillar of strength, was even more devastated, breaking down in tears every time he entered the room. It was the first time I had seen him so vulnerable, and I wanted to comfort him, even as I tried to convince myself that everything would eventually be fine. After a few agonizing days, my mom finally emerged from the coma, marking the happiest moment of my life. From now on, I am determined to show her just how deeply we cherish her. | A couple of months ago, one of the most challenging events I faced nearly happened. My mom was about to undergo surgery due to her weak heart condition, Pulmonary Edema. I was extremely concerned, but I found some comfort in the recent expertise of the doctors working with her. Unfortunately, that assurance wasn't enough. People can make mistakes, and the surgery went wrong. My mom was placed in a coma so doctors could shift from extending her lifespan to actively saving her life. She remained in that state for several days, and I was deeply distraught. My mother was in a critical condition, and I felt helpless. I stayed by her side day and night, unable to eat or sleep, just wishing for her recovery. My dad was even more devastated, unable to enter the room without breaking down. It was the first time I'd seen him so overwhelmed, and I just wanted to comfort him, hoping everything would turn out fine. Ultimately, she pulled through, and it was the happiest moment of my life. From this moment forward, I intend to ensure she knows the depth of our love for her. |